\documentclass[onecolumn]{cas-dc}
\usepackage{float}
\usepackage[numbers]{natbib}
\usepackage{subfigure}

\begin{document}
\let\printorcid\relax
\let\WriteBookmarks\relax
\def\floatpagepagefraction{1}
\def\textpagefraction{.001}
\shorttitle{}
\shortauthors{HeChen Yang et~al.}

\title[mode = title]{EMDS-7: Environmental Microorganism Image Dataset Seventh Version for Multiple Object Detection Evaluation
}  

\author[a]{Hechen Yang}[type=editor,style=chinese,auid=000,bioid=1,]
\author[a]{Chen Li}[style=chinese]
\cormark[1]
\ead{lichen201096@hotmail.com}
\author[b]{Xin Zhao}[style=chinese]
\cormark[2]
\ead{zhaoxin@mail.neu.edu.cn}
\author[a]{Bencheng Cai}[style=chinese]
\author[a]{Jiawei Zhang}[style=chinese]
\author[a]{Pingli Ma}[style=chinese]
\author[a]{Peng Zhao}[style=chinese]
\author[a]{Ao Chen}[style=chinese]
\author[c]{Tao Jiang}[style=chinese]
\author[d]{Hongzan Sun}[style=chinese]
\author[a]{Yueyang Teng}[style=chinese]
\author[a]{Shouliang Qi}[style=chinese]
\author[e]{Xinyu Huang}[style=chinese]
\author[e]{Marcin Grzegorzek}

\address[a]{Microscopic Image and Medical Image Analysis Group, MBIE College, Northeastern University, 110169, Shenyang, PR China}
\cortext[cor1]{Corresponding author}
\address[b]{Department of Environmental Engineering}
\address[c]{School of Control Engineering, Chengdu 
University of Information Technology,Chengdu 610225, China}
\address[d]{China Medical University,China}
\address[e]{Institute of Medical Informatics, University of Luebeck, Luebeck, Germany}

\begin{abstract}
The Environmental Microorganism Image Dataset Seventh Version (EMDS-7) is a microscopic image data set, including the original Environmental Microorganism images (EMs) and the corresponding object labeling files in ".XML" format file. The EMDS-7 data set consists of 41 types of EMs, which has a total of 2365 images and 13216 labeled objects. The EMDS-7 database mainly focuses on the object detection. In order to prove the effectiveness of EMDS-7, we select the most commonly used deep learning methods (Faster-RCNN, YOLOv3, YOLOv4, SSD and RetinaNet) and evaluation indices for testing and evaluation. EMDS-7 is freely published for non-commercial purpose at: https://figshare.com/articles/dataset/EMDS-7\_DataSet/16869571
\end{abstract}


\maketitle

\section{Introduction}

\subsection{Environmental Microorganisms}
\begin{figure*}
	\centering
		\includegraphics[scale=.80]{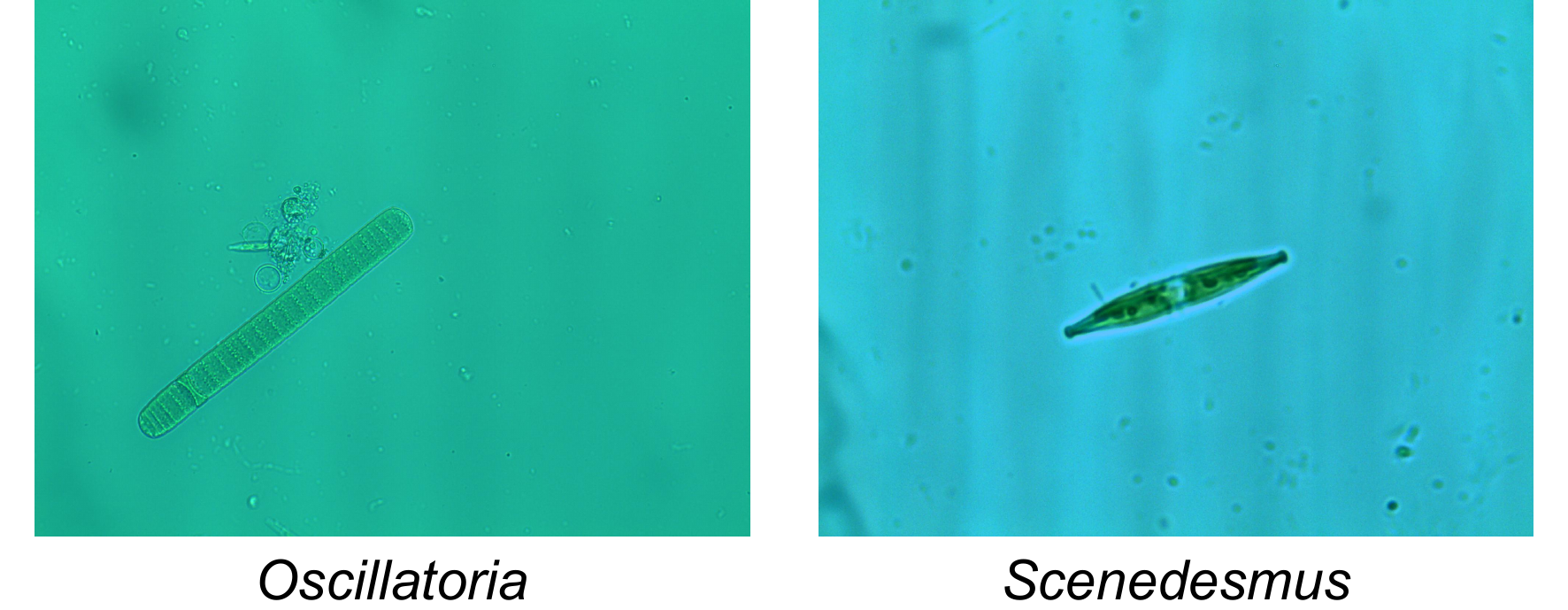}
	\caption{\centering{An example of EM images}}
\label{Fig:EMs}
\end{figure*}

Today, Environmental Microorganisms (EMs) are inseparable from our lives~\cite{Li-2016-EMABC}. 
Some EMs are conducive to the development of ecology and promote the progress of 
human civilization. However, some EMs hinder ecological balance and even cause urban water pollution to affect human health~\cite{anand-2021-sc2a}. Many researchers use microscopes to observe and identify EMs, but sometimes they have deviations in analyzing EMs. 
Computer vision analysis method is of great significance, which can help researchers to analyze the EMs with higher accuracy and efficiency.
For example, \textit{Oscillatoria} is a common EM, which can be observed in various freshwater environments and can thrive in various environments. When it reproduces vigorously, it will produce unpleasant odors, cause water pollution, consume oxygen in the water, and cause fish and shrimp to die of hypoxia~\cite{lu-202-1algicidal}.
In addition, \textit{Scenedesmus} is also a freshwater planktonic algae microorganism, which is usually composed of four to eight cells. \textit{Scenedesmus} has strong resistance to organic pollutants, and plays a vital role in water self-purification and sewage purification~\cite{kashyap-2021-scaao}.   
The images of the EMs proposed above are shown in Fig.~\ref{Fig:EMs}.

\subsection{Application Scenarios of Environmental Microorganism Images}
Image noise appears in the process of acquiring and transmitting EMs images~\cite{Gonzalez-2002-DIP}. Image denoising can reduce the noise of the EM image while preserving the details of the image. Besides, in the process of deep learning based EMs image analyzation, we can extract the features of EM images, then send them to the deep learning network model for training, and match them with known data to classify, retrieve and detect EMs. In addition, EM images can also be applied in the field of image segmentation to separate microorganisms from the complex background of the image~\cite{pal-1993-AROIS}. Meanwhile, EM images can be used in the field of EM object detection. First, we can frame and mark the known EM objects in the original image, and then transfer the image to the object detection model for feature extraction and network training~\cite{zhang-2021-comprehensive}. Finally, the trained model can be applied for object detection of EMs. 

\subsection{Contribution}
EMs are one of the important indicators for investigating the environment, so it is very essential to collect EM data and information~\cite{kosov-2018-EMCUC}. Environmental Microorganism Image Dataset SeventhVersion (EMDS-7) are all taken from urban areas, which can be used to monitor the pollution of the urban water environment. 
Furthermore, due to the constant changes in conditions such as temperature and humidity, EMs are very sensitive to these conditions, so the biomass of EMs are easily affected~\cite{rodriguez2020temporal}. It is difficult to collect enough EM images. Currently, there are some EM data sets, but many of them are not open source. EMDS-7 is provided to researchers as an open source data set. In addition, we prepare high-quality corresponding object label files of EMDS-7 for algorithms and model evaluation. The label file of EMs can be directly used in multiple object detection and analysis. EMDS-7 has a variety of EM images, which provides sufficient data support for EMs object detection and achieves satisfactory detection results.

\section{Dataset Information of EMDS-7}
\begin{figure*}
	\centering
		\includegraphics[scale=.70]{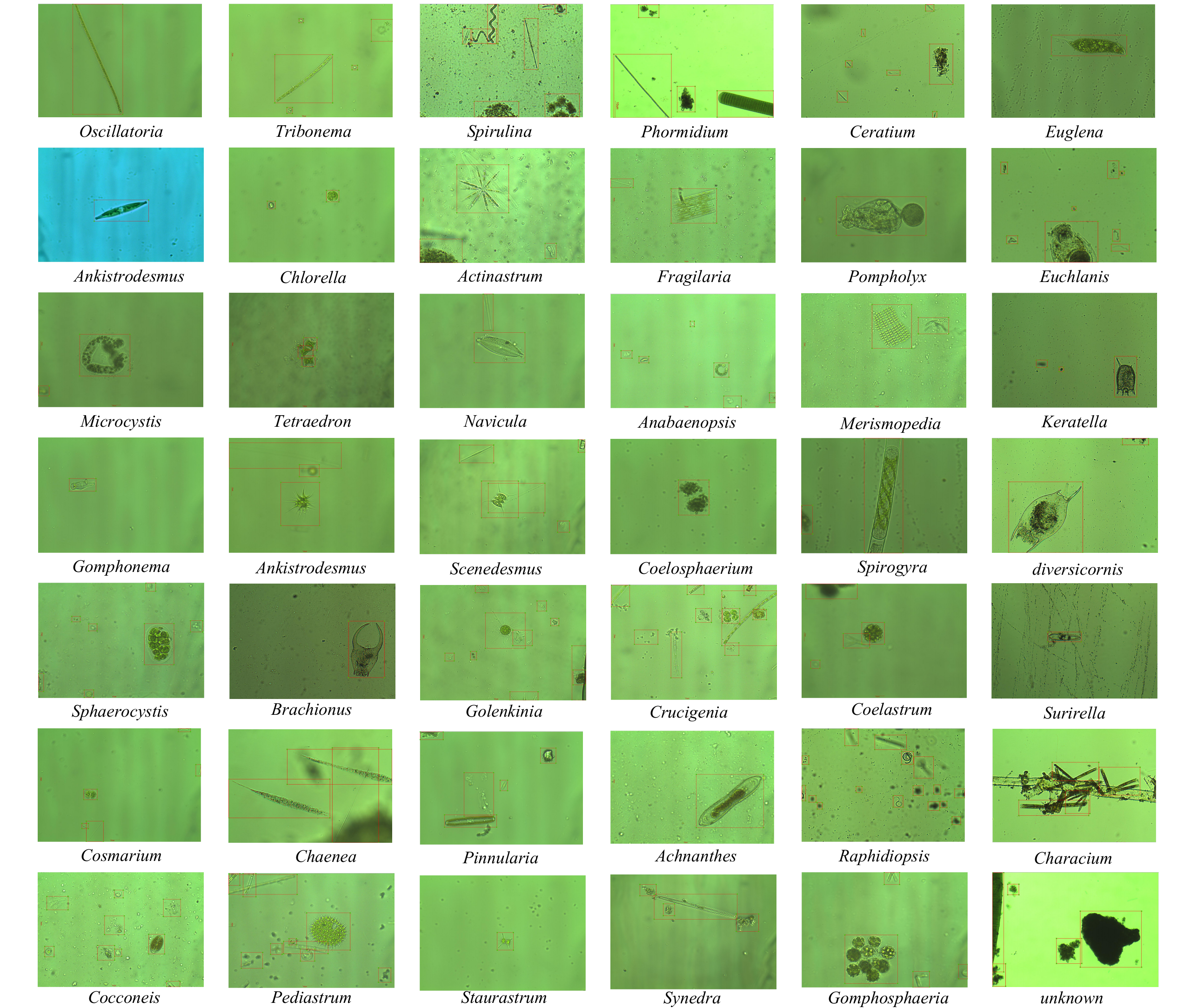}
	\caption{\centering{An example of EM images in EMDS-7.}}\label{Fig:EMDS7}
\end{figure*}

EMDS-7 consists of 2365 images of 42 EM categories and 13216 labeled objects. The EM sampling sources are images taken from different lakes and rivers in Shenyang (Northeast China) by two environmental biologists (Northeastern University, China) under a 400× optical microscope from 2018 to 2019. Then, four bioinformatics scientists (Northeastern University, China) manually prepared the object labeling files in ".XML" format corresponding to the original
2365 images from 2020 to 2021. In the EM object labeling files, 42 types of EMs are labeled by their categories. In addition, the unknown EMs and impurities are marked as Unknown, and a total of 13216 labeled objects are obtained. Table.~\ref{tbl1} shows the basic information of 41 types of EMs and unknown objects; Fig.~\ref{Fig:EMDS7} shows examples of 41 types of EMs and unknown objects in EMDS-7.
The labeled files of EMDS-7 images are manually labeled base on the following two rules:

Rule-A: All identifiable EMs that appear completely or more than 60$\%$ of their own in all images are marked with category labels corresponding to 41 categories.

Rule-B: Unknown EMs that are less than 40$\%$ of their own in all images and EMs other than 41 categories in this database. In addition, some obvious impurities in the background of the image are marked as unknown. 

EMDS-7 is freely published for non-commercial purpose at: https://figshare.com/articles/dataset/EMDS-7\_DataSet/16869571

\begin{table*}[]
\caption{Basic information of 42 EM classes in EMDS-7. Number of original images (NoOI), Number of EM object NEMo), Visible characteristics (VC).}
\label{tbl1}
\resizebox{\textwidth}{40mm}{
\begin{tabular}{|l|l|l|l|l|l|l|l|}

\hline
Classes      & NoOI & NEMo & VC           & Classes & NoOI & NEMo & VC \\ \hline

\textit{Oscillatoria}& 41 & 178  &Cylindrical &\textit{Staurastrum}& 9    & 9    &Multi-radial symmetry     \\ \hline

\textit{Ankistrodesmus}& 5 & 45   &Cell needle &\textit{Phormidium}    & 276  & 1216 &Plant body gelatinous or leathery     \\ \hline
             
\textit{Microcystis}& 307  & 826  &Spherical masses &\textit{Fragilaria} & 55   & 59   &Lanceolate      \\ \hline
             
\textit{Gomphonema} & 87   & 108  &Linear-lanceolate &\textit{Anabaenopsis} & 22   & 37   &Filaments     \\ \hline
             
\textit{Sphaerocystis}& 55   & 53   &Cell sphere &\textit{Coelosphaerium}  & 77   & 165  &Group glue is generous and transparent     \\ \hline
             
\textit{Cosmarium} & 17   & 28   &cell side flattened &\textit{Crucigenia}  & 9    & 16   &Micelles with many cells     \\ \hline
             
\textit{Cocconeis} & 14   & 15   &Flat, oval cells &\textit{Achnanthes} & 18   & 19   &Shell surface linear lanceolate     \\ \hline
             
\textit{Tribonema} & 49   & 88   &Yellow-green cotton-like &\textit{Synedra}& 77   & 206  &Shell needle    \\ \hline
             
\textit{Chlorella} & 80   & 155  &Small round or slightly oval &\textit{Ceratium} & 23   & 24   &Flat back and abdomen     \\ \hline
             
\textit{Tetraedron} & 25   & 66   &Flat or pyramidal &\textit{Pompholyx} & 49   & 51   &Carapace Oval or Shield Shape      \\ \hline
             
\textit{Ankistrodesmus} & 64   & 84   &Needle to spindle &\textit{Merismopedia} & 33   & 38   &Flat group     \\ \hline
             
\textit{Brachionus} & 113  & 144  &quilt is wider and square&\textit{Spirogyra} & 89   & 134  &Strip, spiral     \\ \hline
             
\textit{Chaenea} & 6    & 13   &Cylindrical or spindle &\textit{Coelastrum}& 29   & 30   &Spherical, oval or truncated pyramid     \\ \hline
             
\textit{Pediastrum} & 95   & 105  &Disc or star &\textit{Raphidiopsis}& 9    & 19   &Curved     \\ \hline
             
\textit{Spirulina} & 18   & 73   &Spiral &\textit{Gomphosphaeria}& 58   & 79   &Oval tiny groups     \\ \hline
             
\textit{Actinastrum} & 23   & 181  &Wide round or pointed & \textit{Euglena} & 81   & 81   &Spindle to needle     \\ \hline
             
\textit{Navicula} & 75   & 90   &Shell surface fusiform or oval &\textit{Euchlanis} & 14   & 13   &Oval or pear-shaped     \\ \hline
             
\textit{Scenedesmus} & 86   & 139  &Oval or spindle &\textit{Keratella}& 65   & 69   &Irregular spines     \\ \hline
             
\textit{Golenkinia} & 60   & 279  &Irregularly slender bristles &\textit{diversicornis} & 89   & 99   &Feet have toes and quilt     \\ \hline
             
\textit{Pinnularia} & 36   & 38   &Oval to boat &\textit{Surirella}& 22   & 37   &False shell seam     \\ \hline

\textit{unknown}      &      & 8088 &Unknown EM and impurities &\textit{Characium}& 5    & 19   &Spindle     \\ \hline
\end{tabular}}
\end{table*}

\section{Object Detection Methods for EMDS-7 Evaluation in This Paper}
\subsection{Faster RCNN}
Faster RCNN generates candidate frames based on the Anchor mechanism by adding a region proposal networks (RPN), and finally integrates feature extraction, candidate frame selection, frame regression, and classification into one network, thereby the detection accuracy and efficiency can be effectively improved~\cite{ren-2015-FRTRO}. Faster RCNN achieves high-precision detection performance by adding RPN in two stages, which performs well on multiple data sets. Compared with other one-stage detection networks, the two-stage network is more accurate and can solve the problems of multi-scale and small object detection.
\subsection{YOLOv3}
The object detection model of the YOLO series is a  one-stage detection network, which can locate and classify the objects at the same time. The advantage is that the training speed is fast with less time-consuming. One of the most types is YOLOv3. Joseph Redmon and others used the new basic network darknet-53 in the backbone of YOLOv3 for physical sign extraction~\cite{redmon-2016-yolo}. It contains 53 convolutional layers and introduces a residual structure so that the network can reach a deep level while avoiding the problem of gradient disappearance. In addition, darknet-53 removes the pooling layer and uses a convolutional layer with a step size of 2 to reduce the dimensionality of the feature map, which can maintain the information transmission better. And YOLOv3 also has excellent structures such as anchor and FPN~\cite{lin-2017-FPNFO}.
\subsection{YOLOv4}
YOLOv4 is an improved version based on YOLOv3, which adds CSP and PAN structures~\cite{bochkovskiy-2020-yolov4}. The backbone network of YOLOv3 is modified to CSPDarkne53, and add an spp (spatial pyramid pooling) idea behind the backbone network to expand the receptive field, using 1×1, 5×5, 9×9, 13×13 as the largest Pooling, multi-scale fusion, and improve the accuracy of the model. At the same time, in the neck network of Yolov4, there are Feature Pyramid Network (FPN)~\cite{lin-2017-FPNFO}, Path Aggregation Network (PAN), BiFPN, and NAS-FPN, which can collect different feature maps more effectively. 

\subsection{SSD}
SSD is another striking object detection network after YOLO. SSD has two major advantages~\cite{liu-2016-ssd}. First, SSD extracts feature maps of different scales for detection. Large-scale feature maps can be used to detect small objects, while small-scale feature maps can be used to detect large objects. Second, SSD uses different Prior boxes (Prior boxes, Default boxes, Anchors) for scale and aspect ratio. It follows the method of direct regression box and classification probability in YOLO, and uses anchors to improve recognition accuracy referring to Faster R-CNN. By combining these two networks, SSD balances the advantages and disadvantages of Faster R-CNN and YOLO.
\subsection{RetinaNet}
The RetinaNet object detection model is also a one-stage object detection network. RetinaNet essentially consists of a backbone network (BackBone) and two subnets (SubNet)~\cite{lin-2017-focal}. The backbone network is responsible for calculating the convolution feature map on the entire input image, which is composed of the ResNet residual network and the FPN feature pyramid network. The two sub-networks use the features extracted from the backbone network to achieve their respective functions~\cite{lin-2017-FPNFO}. The first sub-network completes the classification task; the second sub-network completes the bounding box regression task.

\section{Evaluation of Deep Learning Object Detection Methods}
\begin{figure*}
	\centering
		\includegraphics[scale=.50]{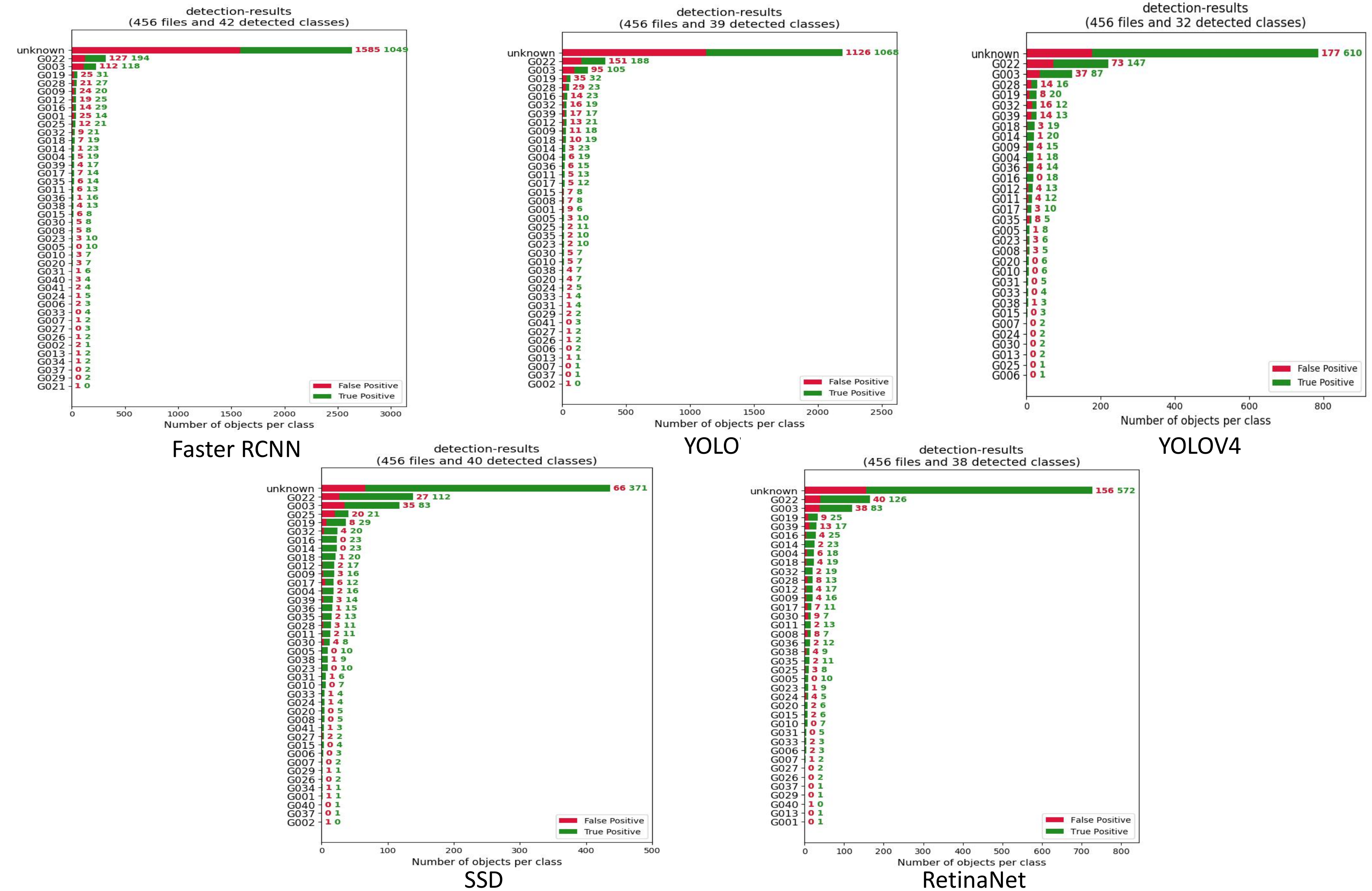}
	\caption{\centering{Five models predict the number of EMDS-7 positive and negative samples.}}\label{Fig:tf}
\end{figure*}

\begin{figure*}
	\centering
		\includegraphics[scale=.50]{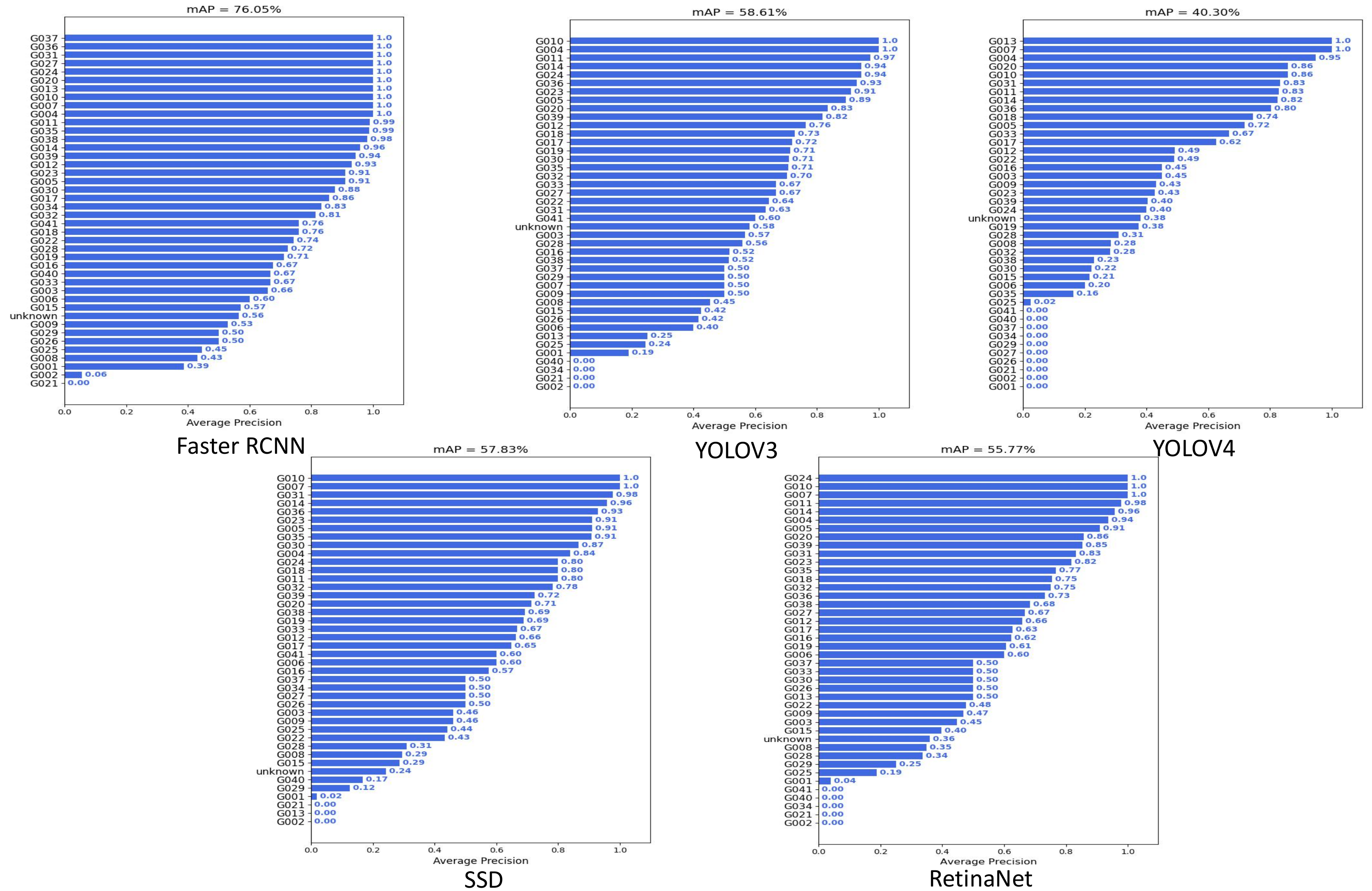}
	\caption{\centering{Each category of EM object detection prediction AP value in EMDS--7}}\label{Fig:ap}
\end{figure*}

\begin{figure*}
	\centering
		\includegraphics[scale=.50]{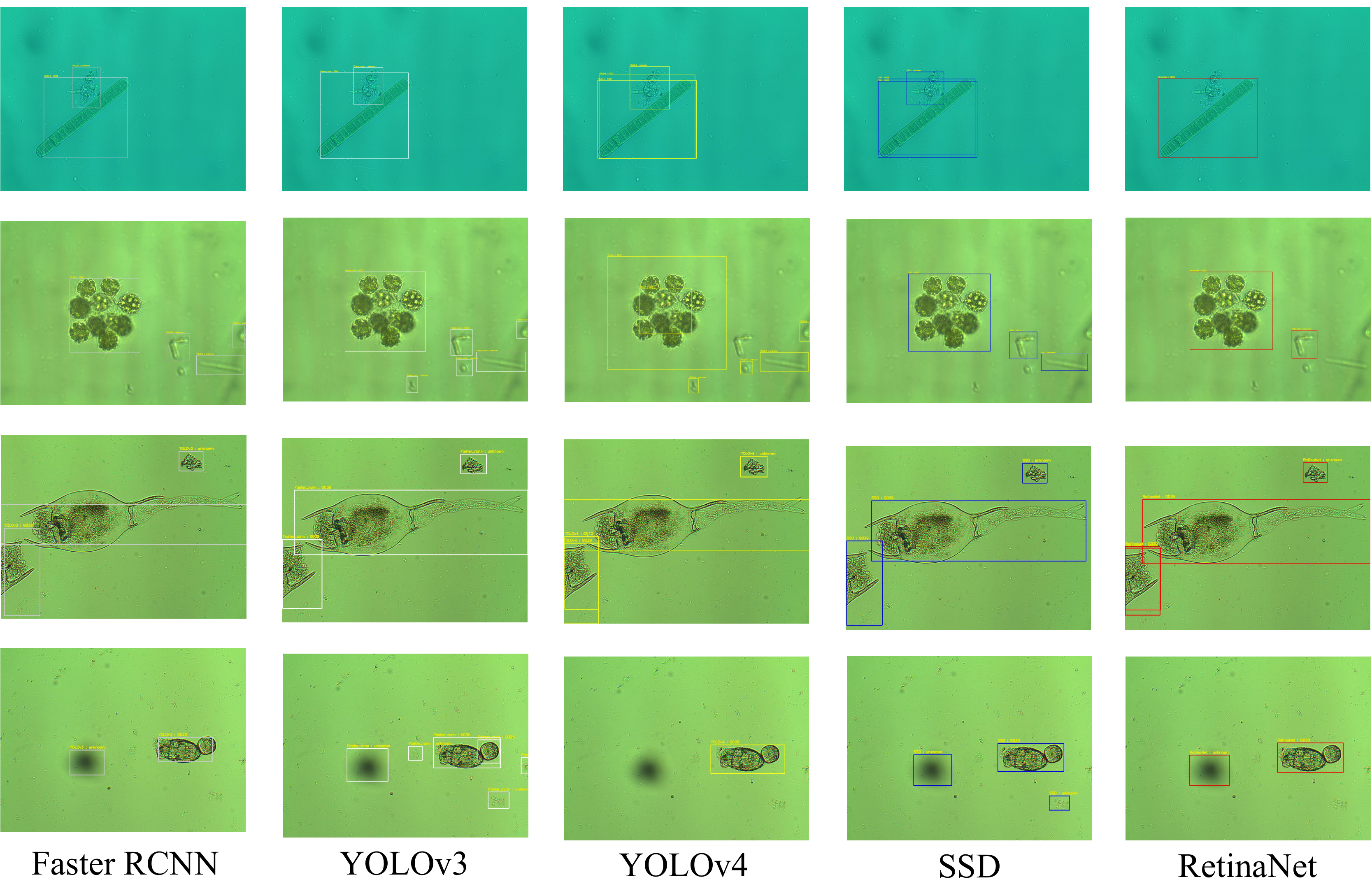}
	\caption{\centering{Five object detection model prediction results in EMDS-7}}\label{Fig:model}
\end{figure*}

Object detection is an important part of image analysis. To prove the effectiveness of EMDS-7 in object detection and evaluation, we use five different deep learning object detection models to detect EMs in the EMDS-7 data set. The five models are Faster RCNN~\cite{ren-2015-FRTRO}, SSD~\cite{liu-2016-ssd}, RetinaNet~\cite{lin-2017-focal}, YOLOv3~\cite{redmon-2016-yolo} and YOLOv4~\cite{bochkovskiy-2020-yolov4}. Because the number of images of each category of EM in the EMDS-7 data set is different, we divide each category of Ems data set into the training, validation and test set according to 6:2:2 to ensure each sets has 42 types of EMs.  We train EMDS-7 for five different kinds of deep learning object detection model, and then respectively predict the images of the test set. We calculate the number of EM objects in 456 EMS images. We set the threshold of the predictive frame confidence to be 0.5, when the model predicts the object's confidence is greater than 0.5, the prediction box is displayed, and Intersection over Union (IOU) is set to 0.3. In Fig.~\ref{Fig:tf} and Fig.~\ref{Fig:ap}, we summarize the number of predictive positive samples and negative samples, and the Average Precision (AP) value of each class of EMs. Analysis of predict results is shown in Tab.~\ref{tbl:3}. We also illustrate the location of the prediction box in the EMS image, and some samples are shown in Fig.\ref{Fig:model}.

In Tab.~\ref{tbl:3} and Fig.~\ref{Fig:ap} we can see that the EMDS-7 data set performs well in the task of object detection, and most of the EMs can be accurately identified. Meanwhile, different object detection models differ in the object detection effect of EMDS-7, which proves that EMDS-7 dataset provide performance analysis of different networks. We also calculate five models of the Mean Average Precision (MAP), which the Faster RCNN value of 76.05$\%$, the YOLOv3 value of 58.61$\%$, the YOLOv4 value of 40.30$\%$, the SSD value of 57.83$\%$, the RetinaNet value of 57.96$\%$. We can see that Faster RCNN has the highest detection performance of EMDS-7. YOLOv3, SSD, RetinaNet model predicts that the MAP value is similar. The lowest is YOLOv4. However, some EMS category predictive AP values can reach 100$\%$, while some kind of AP values are 0$\%$. We found that there are little AP value is 100$\%$ or 0$\%$, and the category of EMs in the image is basically consistent and the EMs characteristics are relatively large. In addition, different models are different from the method of extracting features, so differentiation occurs when model training. For example, in the object detection of Euglena category, the predictive sample has only two, and the AP value is 100$\%$ in the FASTER RCNN, but in the RetinaNET is 0$\%$. In synthesis, the EMDS-7 database can provide analytical performance for different object detection models.

\begin{table*}[]
\caption{A comparison of the object detection  results on test set of EMs.
GTO (Ground Ture Object), tp (ture positive), fp (flase positive) and ap (Average Precision) (In [\%].)}
\label{tbl:3}
\resizebox{\textwidth}{100mm}{
\begin{tabular}{|l|l|l|l|l|l|l|l|l|l|l|l|l|l|l|l|l|}
\hline
               &      & \multicolumn{3}{l|}{Faster RCNN} & \multicolumn{3}{l|}{YOLOv3} & \multicolumn{3}{l|}{YOLOv4} & \multicolumn{3}{l|}{SSD} & \multicolumn{3}{l|}{RetinaNet} \\ \hline
EMS            & GTO  & tp        & fp       & ap        & tp      & fp      & ap      & tp      & fp     & ap       & tp     & fp    & ap      & tp       & fp      & ap        \\ \hline
\textit{Oscillatoria}   & 26   & 14        & 25       & 38.70     & 6       & 9       & 19.08   & 0       & 0      & 0        & 1      & 1     & 1.92    & 1        & 0       & 4.35      \\ \hline
\textit{Ankistrodesmus} & 9    & 1         & 2        & 5.56      & 0       & 1       & 0       & 0       & 0      & 0        & 0      & 1     & 0       & 0        & 0       & 0         \\ \hline
\textit{Microcystis }   & 151  & 118       & 112      & 65.97     & 105     & 95      & 56.66   & 87      & 37     & 44.96    & 83     & 35    & 46.10   & 94       & 31      & 51.61     \\ \hline
\textit{Gomphonema}     & 19   & 19        & 5        & 1.00      & 19      & 6       & 100     & 18      & 1      & 94.74    & 16     & 2     & 83.90   & 18       & 2       & 89.50     \\ \hline
\textit{Sphaerocystis}  & 11   & 10        & 0        & 90.91     & 10      & 3       & 89.26   & 8       & 1      & 71.72    & 10     & 0     & 90.91   & 10       & 0       & 100       \\ \hline
\textit{Cosmarium}      & 5    & 3         & 2        & 60.00     & 2       & 0       & 40.00   & 1       & 0      & 20.00    & 3      & 0     & 60.00   & 3        & 1       & 75.00     \\ \hline
\textit{Cocconeis  }    & 2    & 2         & 1        & 1.00      & 1       & 0       & 50.00   & 2       & 0      & 1.00     & 2      & 0     & 100     & 2        & 0       & 100       \\ \hline
\textit{Tribonema}      & 17   & 8         & 5        & 43.03     & 8       & 7       & 45.45   & 5       & 3      & 28.43    & 5      & 0     & 29.41   & 8        & 7       & 37.11     \\ \hline
\textit{Chlorella}      & 34   & 20        & 24       & 52.82     & 18      & 11      & 49.98   & 15      & 4      & 42.98    & 16     & 3     & 46.05   & 15       & 1       & 48.33     \\ \hline
\textit{Tetraedron}     & 7    & 7         & 3        & 1.00      & 7       & 5       & 1.00    & 6       & 0      & 85.71    & 7      & 0     & 100     & 7        & 1       & 46.67     \\ \hline
\textit{Ankistrodesmus} & 13   & 13        & 6        & 98.97     & 13      & 5       & 97.20   & 12      & 4      & 82.75    & 11     & 2     & 79.94   & 12       & 1       & 98.72     \\ \hline
\textit{Brachionus}     & 25   & 25        & 19       & 93.18     & 21      & 13      & 76.38   & 13      & 4      & 49.10    & 17     & 2     & 66.40   & 25       & 9       & 83.97     \\ \hline
\textit{Chaenea}        & 2    & 2         & 1        & 1.00      & 1       & 1       & 25.00   & 2       & 0      & 100      & 0      & 0     & 0       & 1        & 0       & 50.00     \\ \hline
\textit{Pediastrum}     & 24   & 23        & 1        & 95.83     & 23      & 3       & 94.30   & 20      & 1      & 82.34    & 23     & 0     & 95.83   & 20       & 2       & 95.24     \\ \hline
\textit{Spirulina}      & 14   & 8         & 6        & 57.14     & 8       & 7       & 42.42   & 3       & 0      & 21.43    & 4      & 0     & 28.57   & 9        & 2       & 45.79     \\ \hline
\textit{Actinastrum}    & 40   & 29        & 14       & 67.48     & 23      & 14      & 51.58   & 18      & 0      & 45.00    & 23     & 0     & 57.50   & 29       & 3       & 77.45     \\ \hline
\textit{Navicula}       & 16   & 14        & 7        & 85.83     & 12      & 5       & 71.88   & 10      & 3      & 62.50    & 12     & 6     & 64.91   & 15       & 6       & 72.67     \\ \hline
\textit{Scenedesmus}    & 25   & 19        & 7        & 76.00     & 19      & 10      & 72.88   & 19      & 3      & 74.40    & 20     & 1     & 80.00   & 18       & 0       & 75.00     \\ \hline
\textit{Golenkinia}     & 41   & 31        & 25       & 71.15     & 32      & 35      & 71.30   & 20      & 8      & 37.52    & 29     & 8     & 68.88   & 29       & 9       & 60.57     \\ \hline
\textit{Pinnularia}     & 7    & 7         & 3        & 1.00      & 7       & 4       & 83.47   & 6       & 0      & 85.71    & 5      & 0     & 71.43   & 6        & 1       & 75.00     \\ \hline
\textit{Staurastrum}    & 3    & 0         & 1        & 0         & 0       & 0       & 0       & 1       & 1      & 0        & 0      & 0     & 0       & 0        & 0       & 0         \\ \hline
\textit{Phormidium}     & 234  & 194       & 127      & 74.23     & 188     & 151     & 64.49   & 147     & 73     & 48.84    & 4      & 1     & 43.19   & 142      & 32      & 52.93     \\ \hline
\textit{Fragilaria}     & 11   & 10        & 3        & 90.91     & 10      & 2       & 90.91   & 6       & 3      & 42.73    & 21     & 20    & 90.91   & 10       & 3       & 90.91     \\ \hline
\textit{Anabaenopsis}   & 5    & 5         & 1        & 1.00      & 5       & 2       & 94.29   & 2       & 0      & 40.00    & 2      & 0     & 80.00   & 9        & 2       & 81.82     \\ \hline
\textit{Coelosphaerium} & 42   & 21        & 12       & 44.58     & 11      & 2       & 24.43   & 1       & 0      & 2.38     & 2      & 2     & 44.19   & 6        & 3       & 18.30     \\ \hline
\textit{Crucigenia}     & 4    & 2         & 1        & 50.00     & 2       & 1       & 41.67   & 0       & 0      & 0        & 11     & 3     & 50.00   & 2        & 0       & 66.67     \\ \hline
\textit{Achnanthes}     & 3    & 3         & 0        & 1.00      & 2       & 1       & 66.67   & 0       & 0      & 0        & 2      & 2     & 50.00   & 3        & 0       & 100       \\ \hline
\textit{Synedra}        & 34   & 27        & 21       & 72.42     & 23      & 29      & 55.90   & 16      & 14     & 30.92    & 11     & 3     & 31.00   & 15       & 1       & 34.09     \\ \hline
\textit{Ceratium}       & 4    & 2         & 0        & 50.00     & 2       & 2       & 50.00   & 0       & 0      & 0        & 1      & 1     & 12.50   & 1        & 1       & 33.33     \\ \hline
\textit{Pompholyx}      & 9    & 8         & 5        & 87.65     & 7       & 5       & 70.83   & 2       & 0      & 22.22    & 8      & 4     & 86.67   & 8        & 3       & 85.28     \\ \hline
\textit{Merismopedia}   & 6    & 6         & 1        & 1.00      & 4       & 1       & 63.33   & 5       & 0      & 83.33    & 6      & 1     & 97.62   & 5        & 0       & 71.43     \\ \hline
\textit{Spirogyra}      & 25   & 21        & 9        & 81.31     & 19      & 16      & 70.23   & 12      & 16     & 28.14    & 20     & 4     & 78.18   & 21       & 2       & 73.50     \\ \hline
\textit{Coelastrum}     & 6    & 4         & 0        & 66.67     & 4       & 1       & 66.67   & 4       & 0      & 66.67    & 4      & 1     & 66.67   & 3        & 0       & 42.86     \\ \hline
\textit{Raphidiopsis}   & 2    & 2         & 1        & 83.33     & 0       & 0       & 0       & 0       & 0      & 0        & 1      & 1     & 50.00   & 0        & 0       & 0         \\ \hline
\textit{Gomphosphaeria} & 14   & 14        & 6        & 98.74     & 10      & 2       & 70.78   & 5       & 8      & 16.23    & 13     & 2     & 90.81   & 11       & 3       & 100       \\ \hline
\textit{Euglena}        & 16   & 16        & 1        & 1.00      & 15      & 6       & 92.97   & 14      & 4      & 80.28    & 15     & 1     & 92.97   & 12       & 1       & 73.56     \\ \hline
\textit{Euchlanis}      & 2    & 2         & 0        & 1.00      & 1       & 0       & 50.00   & 0       & 0      & 0        & 1      & 0     & 50.00   & 0        & 0       & 0         \\ \hline
\textit{Keratella}      & 13   & 13        & 4        & 98.19     & 7       & 4       & 51.54   & 3       & 1      & 23.08    & 9      & 1     & 69.23   & 9        & 1       & 69.23     \\ \hline
\textit{diversicornis}  & 18   & 17        & 4        & 94.44     & 17      & 17      & 81.71   & 13      & 14     & 40.30    & 14     & 3     & 72.47   & 19       & 11      & 92.78     \\ \hline
\textit{Surirella}     & 6    & 4         & 3        & 66.67     & 0       & 0       & 0       & 0       & 0      & 0        & 1      & 0     & 16.67   & 1        & 1       & 16.67     \\ \hline
\textit{Characium}      & 5    & 4         & 2        & 76.00     & 3       & 0       & 60.00   & 0       & 0      & 0        & 3      & 1     & 60.00   & 0        & 0       & 0         \\ \hline
\textit{unknown}        & 1429 & 1049      & 1585     & 56.49     & 1068    & 1126    & 58.17   & 610     & 177    & 38.15    & 371    & 66    & 24.17   & 622      & 146     & 36.92     \\ \hline
\end{tabular}}
\end{table*}

\section{Conclusion and Future Work}
EMDS-7 is an object detection data set containing 42 types of EMs, which contains the original image of EMs and object label data for corresponding EMs. EMDS-7 labeled 15342 EMs. At the same time, we further add some deep learning object detection experiments to the EMDS-7 database to prove the effectiveness.
During the object detection process, we divide the data set according to 6: 2: 2 for train, validation and test sets. We use five different deep learning object detection methods to test EMDS-7 and use multiple evaluation indices to evaluate the prediction results.

In the future, we will enlarge the category of EMs to increase the number of images of each EM. Make each class data balanced and sufficient. We hope to use the EMDS-7 database to achieve more features in the future. Meanwhile

\section*{Acknowledgements}
This work is supported by the ``National Natural Science Foundation of China'' (No.61806047) 
and the ``Fundamental Research Funds for the Central Universities'' (No. N2019003). 
We thank Miss Zixian Li and Mr. Guoxian Li for their important discussion.

\section*{Declaration of competing interest}
The authors declare that they have no conflict of interest.
\bibliographystyle{unsrt}
\bibliography{wenxian}
\end{document}